\RequirePackage{amsmath}
\documentclass[runningheads]{llncs}
\usepackage[british]{babel}
\usepackage[caption=false]{subfig}
\usepackage{graphicx}
\usepackage{color}
\usepackage{xcolor}
\usepackage{amsfonts}
\usepackage{amsmath}
\usepackage{mathtools}
\usepackage{wrapfig}
\usepackage{graphicx}
\usepackage{svg}
\usepackage{subfig}
\usepackage{dsfont}
\usepackage{amsmath}
\usepackage{tikz}
\usepackage{mathdots}

\usepackage{cancel}
\usepackage{color}
\usepackage{siunitx}
\usepackage{array}
\usepackage{multirow}
\usepackage{amssymb}
\usepackage{tabularx}
\usepackage{booktabs}
\usepackage{array}
\usepackage{makecell}

\usetikzlibrary{fadings}
\usetikzlibrary{patterns}

\begin{document}
\title{Increasing negotiation performance at the edge of the network\thanks{This research was sponsored by the U.S. Army Research Laboratory and the U.K. Ministry of Defence under Agreement Number W911NF-16-3-0001. The views and conclusions contained in this document are those of the authors and should not be interpreted as representing the official policies, either expressed or implied, of the U.S. Army Research Laboratory, the U.S. Government, the U.K. Ministry of Defence or the U.K. Government. The U.S. and U.K. Governments are authorized to reproduce and distribute reprints for Government purposes notwithstanding any copyright notation hereon. }}
\author{Sam Vente\inst{1}\orcidID{0000-0002-5485-6571} \and
Angelika Kimmig\inst{1}\orcidID{0000-0002-6742-4057} \and
Alun Preece\inst{1}\orcidID{0000-0003-0349-9057} \and
Federico Cerutti \inst{2}~ \inst{1}\orcidID{0000-0003-0755-0358}}
\authorrunning{S. Vente et. al.}
\institute{Cardiff University, Cardiff, CF10 3AT, UK 
\email{\{VenteDA,KimmigA,PreeceAD,CeruttiF\}@cardiff.ac.uk} \and
Department of Information Engineering, University of Brescia, Italy \email{federico.cerutti@unibs.it}}

\maketitle              % typeset the header of the contribution

\begin{abstract}  
Automated negotiation has been used in a variety of distributed settings, such as privacy in the Internet of Things (IoT) devices and power distribution in Smart Grids. The most common protocol under which these agents negotiate is the Alternating Offers Protocol (AOP). Under this protocol, agents cannot express any additional information to each other besides a counter offer. This can lead to unnecessarily long negotiations when, for example, negotiations are impossible, risking to waste bandwidth that is a precious resource at the edge of the network. While alternative protocols exist which alleviate this problem, these solutions are too complex for low power devices, such as IoT sensors operating at the edge of the network. To improve this bottleneck, we introduce an extension to AOP called Alternating Constrained Offers Protocol (ACOP), in which agents can also express constraints to each other. This allows agents to both search the possibility space more efficiently and recognise impossible situations sooner. We empirically show that agents using ACOP can significantly reduce the number of messages a negotiation takes, independently of the strategy agents choose. In particular, we show our method significantly reduces the number of messages when an agreement is not possible. Furthermore, when an agreement is possible it reaches this agreement sooner with no negative effect on the utility.
\end{abstract}

\keywords{Automated negotiation\and Multi-agent systems\and Constraints} 

\section{Introduction}
Autonomous agents, in particular those at the edge of the network---near to the source of the data like single or cooperative sensors---often need to coordinate actions to achieve a shared goal: for instance, they might need to negotiate either access to local data to learn a shared model; or access to a shared resource like bandwidth; or joint actions for complex activities such as patrolling an area against wildfires (cf. Section \ref{ref:context}).

Automated negotiation can provide a solution, by allowing agents to reach a mutual consensus on what should and what should not be shared. However, the standard method of negotiation under the Alternating Offers Protocol (AOP, Section \ref{sec:aop})
\cite{Aydogan2017}
can be resource intensive and in particular bandwidth intensive due to the number of messages that need to be exchanged before an outcome can be determined. This might be particularly wasteful when considering autonomous agents at the edge of the network, which have limited bandwidth resources. At the same time, because such agents are often deployed on low-power devices, they cannot be equipped with extremely complex reasoning capabilities able to learn and predict other agents' behaviour.

In Section \ref{sec:acop} we present a novel extension of AOP called Alternating Constrained Offers Protocol (ACOP), that provides a suitable trade-off between reasoning capabilities and bandwidth usage, allowing agents to express constraints on any possible solution along with the proposals they generate. This allows agents to search more effectively for proposals that have a higher probability of being accepted by the adversary. To measure the impact of this on the length and outcomes of negotiations, we perform empirical analysis on a dataset of simulated negotiations (Section \ref{sec:methodology}). To summarise, in this work we will address the following questions:
\begin{enumerate}
  \item[Q1] Do negotiations operating under ACOP exchange fewer messages than negotiations operating under AOP in similar scenarios?
  \item[Q2] Does adopting ACOP negatively impact the outcome of negotiations when compared to negotiations using AOP?
\end{enumerate}

Results summarised in Section \ref{sec:results} provide evidence that negotiations operating under ACOP require substantially fewer messages than negotiations operating under AOP, without negatively affecting the utility of the outcome.

\section{Context and Motivating Examples}
\label{ref:context}

Automated negotiation \cite{fatima2014principles} is a wide field: while our focus is much narrower, it encompasses a substantial number of application domains such as but not limited to, resource allocation, traffic flow direction, e-commerce, and directing Unmanned Vehicles (UxVs) \cite{yassineDoubleAuctionMechanisms2019,lamparterAgentbasedMarketPlatform2010,Zhang2010}. As mentioned before, the most commonly used protocol for automated negotiation, AOP, can require large amounts of messages to be communicated before an outcome can be determined. While more sophisticated methods that alleviate communication bottlenecks by using, for example, fully-fledged constraint satisfaction solvers \cite{gaciarzConstraintBasedNegotiationModel2015} exist, these can include very complex reasoning that is not appropriate for agents deployed on low-power devices that operate at the edge of a network. Additionally, many of these solutions require a neutral third party to act as a mediator, which is not always possible in distributed or adversarial settings. Below we will explore three examples to illustrate some of these applications.

Firstly, autonomous agents can share the burden of learning a model. Federated Learning is a machine learning setting where the goal is to train a high-quality model with training data distributed over a large number of agents, each possibly with unreliable and relatively slow network connections \cite{45648} and with constraints such as limited battery power. For instance, in \cite{Xu:2019:EFL:3321408.3323080} the authors introduce an incentive mechanism using auction-like strategies to
negotiate with bidding in a format similar to \cite{8462804}.

A second domain concerns the negotiation of wireless spectrum allocation \cite{8462804,delaHoz2015}. For instance, due to the low cost of IP-based cameras, wireless surveillance sensor networks are now able to monitor large areas. 
These networks thus require frequency channels to be assigned in a clever way: to this end, in \cite{delaHoz2015} the authors propose to use a text mediation protocol \cite{klein2003negotiating}.

Consider now our third case, that involves a fully distributed and autonomous surveillance system such as using Unmanned Aerial Vehicles (UAVs) to patrol an area at high risk of wildfires. Each UAV is fully autonomous and equipped with processing capability for analysing their sensor streams and detect early signs of wildfire. The uplink to the command control centre is via a slow and unreliable satellite connection. However, each UAV is aware of the existence of other UAVs via low-bandwidth wireless connections. Each UAV has access to commercial-grade GPS. All UAVs are programmed to jointly cover a given area, and have access to high-quality maps of the area which includes detailed level curves. For simplicity, let us assume that the area is divided into sectors, and each UAV announces the sector where it is, and the sector where it intends to proceed. 

Each UAV begins its mission randomly choosing a direction, and hence the next sector it will visit. Its main goal is to preserve its own integrity---after all it is worth several hundreds thousands dollars---while collaborating towards the achievement of the shared goal. It is therefore allowed to return to base, even if this will entail that the shared goal will not be achieved. Examples of this include, when its battery cell level is too low,  when adverse weather conditions affect the efficiency of the UAV rotors, or when it has been damaged by in-flight collision or some other unpredictable situation. In the case two UAVs announce that they are moving towards the same sector, a negotiation between them needs to take place in order to achieve coverage of the sector, while avoiding unnecessary report duplication. 

Let us suppose UAV1 receives an update that UAV2 can visit sector Sierra, the same sector it was also aiming at. It can then send a negotiation offer to UAV2 asking to be responsible for Sierra. UAV2 most likely will at first reply that it should take care of Sierra, while UAV1 can take care of the nearby Tango: after all, it announced it first, it is already en route, and it needs to protect its own integrity. Let us suppose that UAV1 knows that with its current power level and/or performance of its 18 rotors, it cannot visit sector Tango as it would require a substantial lifting. It would then be useful for it to communicate such a constraint, so to shorten the negotiation phase and proceed towards an agreement (or a certification of a disagreement) in a short time frame. Indeed, knowing of UAV1's constraint, UAV2 can accept to visit Tango, or maybe not, due to other constraints. In the latter case, UAV1 can then quickly proceed to search for other sectors to visit, or, alternatively, to return to base.

This last example illustrates potential uses of being able to communicate constraints to other agents. In the next section we will set up the necessary theory to discuss our proposed solution.

\section{Background in Alternating Offers Protocol}
\label{sec:aop}
Firstly we will give a brief overview of the basic negotiation theory used in this work. Here all negotiations are assumed to be \textit{bilateral}, meaning between only two agents, referred to as $A$ and $B$ respectively. The \textit{negotiation space}, which is denoted $\Omega$, represents the space of allowable proposals. This consists of the product of several sets called \textit{issues}, each containing a finite number of elements called \textit{values}. So, to reiterate, when we write $\Omega = \prod_{i=0}^N \Lambda_i$ with $|\Lambda_i| = M_i$ that means that the negotiation consists of $N$ issues consisting of $M_i$ values. In the case that $\forall i,j \in \{1,\ldots,N\}: \Lambda_i = \Lambda_j$ we may also write $\Omega = \Lambda ^N$. Each agent is also assumed to have a \textit{utility function} $u_A, u_B: \Omega \to \mathbb{R}$ which each induce a total preorder $\succeq_A$ and $\succeq_B$ on $\Omega$ via the following relation $$\forall \omega, \zeta \in \Omega: \omega \succeq_A \zeta \iff u_A(\omega) \geq u_A(\zeta)$$ and analogous for $B$, allowing the agents to decide whether they prefer one proposal to another, vice versa or are indifferent towards them. Each agent also has a \textit{reservation value} $\rho_A,\rho_B$ respectively, which is the minimum utility an offer must have to an agent to be acceptable.  A utility function $u$ is called \textit{linearly additive} when the following identity holds: \begin{equation}
  \forall \omega \in \Omega: u(\omega) = \sum_{i=1}^n w_i e_i(\omega_i) \label{linAdd}
\end{equation}
Here $\sum_{i=0}^N w_i = 1$ and $\forall i \in \{1,\ldots,N\} : w_i \in [0,1]$. Here the $w_i$ represents the relative importance of the $i$th issue. This makes explicit that the assignment of any issue does not influence the utility of any of the other issues.

The way in which the agents communicate is detailed by the \textit{protocol}. This is a technical specification of the modes of communication and what types of communication are allowed. The most commonly used protocol is called the Alternating Offers Protocol (AOP). In this protocol, the agents have only three options: make a proposal, accept the previous proposal or terminate the interaction without coming to an agreement. Here we use $\omega^t$ to denote the offer made at time-step $t$. Note that $t$ is discrete.

Finally, agents explore the negotiation space according to their \textit{strategy}. Two well known examples, known as \textit{zero intelligence} and \textit{concession} \cite{baarslagLearningOpponentAutomated2016}. The zero intelligence strategy is also referred to as a \textit{random sampling} strategy. Agents using a random sampling strategy generate offers by simply defining a uniform distribution over the values of each issue, and constructs offers by sampling from those distributions until they find one that is acceptable to them. Agents using a concession strategy might just simply enumerate the offers in the negotiation space in descending order of preference,  either until the other accepts or until they are unable to find offers that they find acceptable. We will use these two strategies in our empirical analysis below. Both these strategies are well known in the literature \cite{lopesConcessionStrategiesNegotiating2012a,azarNonCooperativeFrameworkCoordinating2019,yangMultidemandNegotiationModel2019,godeAllocativeEfficiencyMarkets1993,baarslagAcceptingOptimallyAutomated2013a,wangIntelligentNegotiationAgent2012,hindriksUsingOpponentModels2009}. Zero Intelligence agents are often used as a baseline for benchmarks and concession strategies in various forms are well studied \cite{baarslagLearningOpponentAutomated2016}. We therefore use them here as a proof of concept.

\section{Our Proposal: Alternating Constrained Offers Protocol}
\label{sec:acop}
Almost any negotiation is subject to certain constraints. For example, a good faith agent will never be able to agree to sell something they do not have. When constraints are incompatible, this can dramatically increase the length of the negotiation, since under AOP there is no way to communicate boundaries of acceptable offers. In an effort to alleviate this problem, without introducing too much complexity, we propose an extension of AOP called Alternating Constrained Offers Protocol (ACOP). Using this protocol agents have the opportunity to express a constraint to the opponent when they propose a counter offer. This constraint makes evident that any proposal not satisfying this constraint will be rejected apriori.

In this way, agents can express more information to the opponent about which part of the negotiation space would be useful to explore without having to reveal too much information about their utility function. This can even present some strategic options. Cooperative agents could express all their constraints as fast as possible to give the opponent more information to come up with efficient proposals. On the other hand, more conservative agents can express constraints only as they become relevant, which might lead to expose fewer information in the case the negotiation terminates with an agreement before exposing \emph{red lines}. In this work we focus on the use of \textit{atomic constraints}. These are  constraints that express which one of  single particular issue value assignments is unacceptable. These constraints can either be given to the agent apriori, or they can be deduced by the agent themselves. Especially in the discrete case with linear utility functions, a simple branch and bound search algorithm can be enough to deduce where certain constraints can be created, which we illustrate with the following example.

\begin{figure}[ht]
  \centering
  \subfloat[A's utility matrix]{\includegraphics[width=0.35\columnwidth]{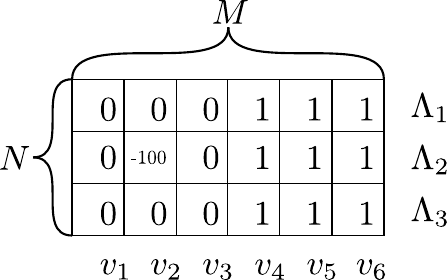}}
  \subfloat[B's utility matrix]{\includegraphics[width=0.35\columnwidth]{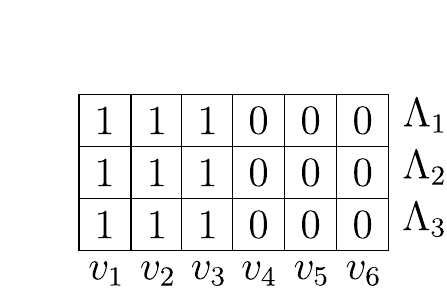}}
  \caption{Utility matrices for A and B respectively for example \ref{ConstraintGenExample}}
  \label{utilMatrices}
\end{figure}

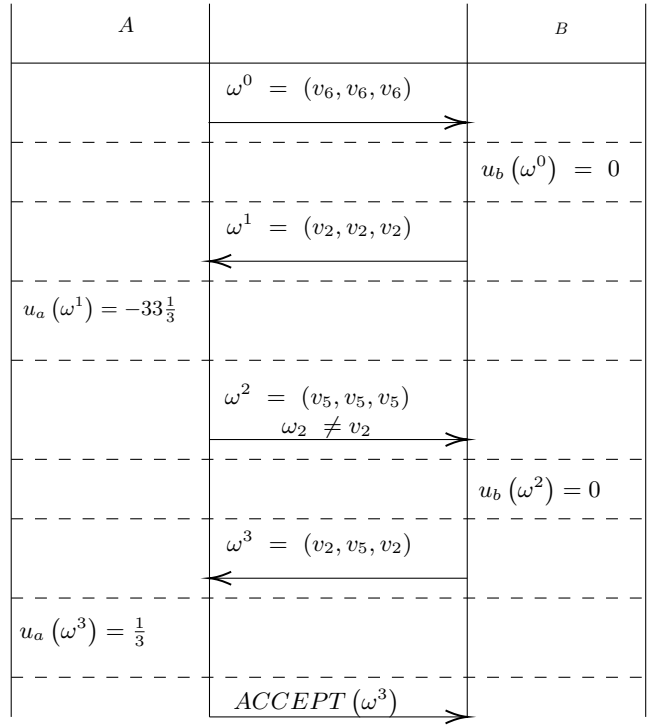
\begin{figure}[ht]
  \centering
  % tikz code generated by Matcha.io
  \begin{tikzpicture}[x=0.75pt,y=0.75pt,yscale=-1,xscale=1]
    \draw    (10,40) -- (330,40) ;
    \draw    (110,10) -- (110,370) ;
    \draw    (240,10) -- (240,370) ;
    \draw    (330,10) -- (330,370) ;
    \draw    (10,10) -- (10,370) ;
    \draw    (109.5,70) -- (210,70) -- (237.5,70) ;
    \draw [shift={(239.5,70)}, rotate = 180] [color={rgb, 255:red, 0; green, 0; blue, 0 }  ][line width=0.75]    (10.93,-3.29) .. controls (6.95,-1.4) and (3.31,-0.3) .. (0,0) .. controls (3.31,0.3) and (6.95,1.4) .. (10.93,3.29)   ;
    \draw    (240,140) -- (112,140) ;
    \draw [shift={(110,140)}, rotate = 360] [color={rgb, 255:red, 0; green, 0; blue, 0 }  ][line width=0.75]    (10.93,-3.29) .. controls (6.95,-1.4) and (3.31,-0.3) .. (0,0) .. controls (3.31,0.3) and (6.95,1.4) .. (10.93,3.29)   ;
    \draw    (110,230) -- (238,230) ;
    \draw [shift={(240,230)}, rotate = 180] [color={rgb, 255:red, 0; green, 0; blue, 0 }  ][line width=0.75]    (10.93,-3.29) .. controls (6.95,-1.4) and (3.31,-0.3) .. (0,0) .. controls (3.31,0.3) and (6.95,1.4) .. (10.93,3.29)   ;
    \draw    (240,300) -- (223.5,300) -- (112,300) ;
    \draw [shift={(110,300)}, rotate = 360] [color={rgb, 255:red, 0; green, 0; blue, 0 }  ][line width=0.75]    (10.93,-3.29) .. controls (6.95,-1.4) and (3.31,-0.3) .. (0,0) .. controls (3.31,0.3) and (6.95,1.4) .. (10.93,3.29)   ;
    \draw    (110,370) -- (152.5,370) -- (238,370) ;
    \draw [shift={(240,370)}, rotate = 180] [color={rgb, 255:red, 0; green, 0; blue, 0 }  ][line width=0.75]    (10.93,-3.29) .. controls (6.95,-1.4) and (3.31,-0.3) .. (0,0) .. controls (3.31,0.3) and (6.95,1.4) .. (10.93,3.29)   ;
    \draw  [dash pattern={on 4.5pt off 4.5pt}]  (10,80) -- (330,80) ;
    \draw  [dash pattern={on 4.5pt off 4.5pt}]  (10,110) -- (330,110) ;
    \draw  [dash pattern={on 4.5pt off 4.5pt}]  (10,150) -- (201.5,150) -- (330,150) ;
    \draw  [dash pattern={on 4.5pt off 4.5pt}]  (10,190) -- (330,190) ;
    \draw  [dash pattern={on 4.5pt off 4.5pt}]  (10,240) -- (197.5,240) -- (330,240) ;
    \draw  [dash pattern={on 4.5pt off 4.5pt}]  (10,270) -- (295.5,270) -- (330,270) ;
    \draw  [dash pattern={on 4.5pt off 4.5pt}]  (10,310) -- (330,310) ;
    \draw  [dash pattern={on 4.5pt off 4.5pt}]  (10,350) -- (288.5,350) -- (330,350) ;
    \draw (68,20.5) node [scale=0.9]  {$A$};
    \draw (165.5,52.5) node   {$\omega ^{0} \ =\ ( v_{6} ,v_{6} ,v_{6})$};
    \draw (287.5,22.5) node [scale=0.7]  {$B$};
    \draw (282,94) node   {$u_{b}\left( \omega ^{0}\right) \ =\ 0$};
    \draw (165.5,122.5) node   {$\omega ^{1} \ =\ ( v_{2} ,v_{2} ,v_{2})$};
    \draw (55,164.5) node [scale=0.9]  {$u_{a}\left( \omega ^{1}\right) =-33\frac{1}{3}$};
    \draw (164.5,216.5) node   {$ \begin{array}{l}
          \omega ^{2} \ =\ ( v_{5} ,v_{5} ,v_{5}) \\
          \ \ \ \ \ \ \ \omega _{2} \ \neq v_{2}
        \end{array}$};
    \draw (276.5,256) node   {$u_{b}\left( \omega ^{2}\right) =0$};
    \draw (165.5,282.5) node   {$\omega ^{3} \ =\ ( v_{2} ,v_{5} ,v_{2})$};
    \draw (46.5,327) node   {$u_{a}\left( \omega ^{3}\right) =\frac{1}{3}$};
    \draw (164,363) node   {$ACCEPT\left( \omega ^{3}\right)$};
  \end{tikzpicture}

  \caption{A schematic representation of an example negotiation under ACOP in the setting set out in Example \ref{ConstraintGenExample} assuming both agents use uniform weights}
  \label{fullACOPExample}
\end{figure}

\begin{example}\label{ConstraintGenExample}
  Let $A,B$ both be negotiation agents having the reservation value $\frac 13$ and linear additive utility functions $u_A, u_B$ respectively, using uniform importance weights. Furthermore, let $\Omega = \Lambda^3$ with $\Lambda = \{v_1,\ldots,v_6\}$. Therefore we have 3 issues, with 6 values each. In this setup we can represent $u_A$ and $u_B$ as matrices which are depicted in Figure \ref{utilMatrices}, with the rows representing the issues and the columns possible values. For example the offer $\omega= (v_1,v_1,v_1)$ would have 0 utility for $A$ and thus be unacceptable but utility 1 for $B$ and be acceptable. Due to the scale of the potential losses $A$ can deduce using branch and bound that $\omega_2 = v_2$ can never be part of a solution they could accept. Therefore they can record this constraint, and express this to $B$ according to their strategy. An example of a negotiation under ACOP of this scenario can be seen in Figure~\ref{fullACOPExample}.
\end{example}

This kind of reasoning is simple enough that it could be evaluated in response to new information, such as an opponent ruling out a crucial option during a negotiation. These constraints can help agents find acceptable options more efficiently, but are also useful to help agents terminate faster by letting them realise that a negotiation has no chance of succeeding. For example, when each possible value of a particular issue is ruled out by at least one of the participants, agreement is impossible and the agents can terminate early.

\section{Experimental methodology}
\label{sec:methodology}
Our empirical analysis provides evidence that ACOP improves over AOP in terms of negotiation length and does not negatively impact utility. We simulated a variety of negotiations with randomly generated problems and  agents using either a random sampling or concession strategy as defined earlier, both under AOP and ACOP. At the end of a simulation we recorded metrics such as length of the negotiation and the outcome. In this section we will first detail how the problems were generated and how the simulations were run. Then we will discuss the results in more detail in the next section.

\subsection{Problem generation}
To run a simulation of a negotiation, four things are required:
\begin{enumerate}
  \item A negotiation space.
  \item The utility functions for the two agents.
  \item The reservation value for both agents.
  \item The strategy and protocol the agents will use (in this case they are always equal for both agents).
\end{enumerate}

To make the results easier to compare, the negotiation space remained constant, consisting of 5 issues each with 5 values across all negotiations. The utility function and the reservation value determine which part of the negotiation space is acceptable to which of the agents, whereas an agent's strategy determines how they explore the possibility space. We refer to an offer which is acceptable to both participants of a negotiation as a \textit{solution} to that negotiation. Furthermore we call a negotiation \textit{possible} if there exists at least one solution, and otherwise \textit{impossible}. We use \textit{configuration} to refer to a pair of utility functions and a pair of reservation values. A pair of utility functions is referred to as a \textit{scenario}. Note that for any configuration, the number of solutions can be calculated to any outside observer with perfect information, since this is deterministic given the parameters. In total $261,225$ configurations were generated, for each of which 4 negotiations were simulated, each corresponding to one of the strategy and protocol pairs. This means that in total $1,044,900$ negotiations were simulated and for each of them the length of the negotiation and the utility the agents achieved at the end were recorded. 

\begin{figure}
  \centering
  \subfloat[Scenario as originally generated]{\includegraphics[width=0.5\columnwidth]{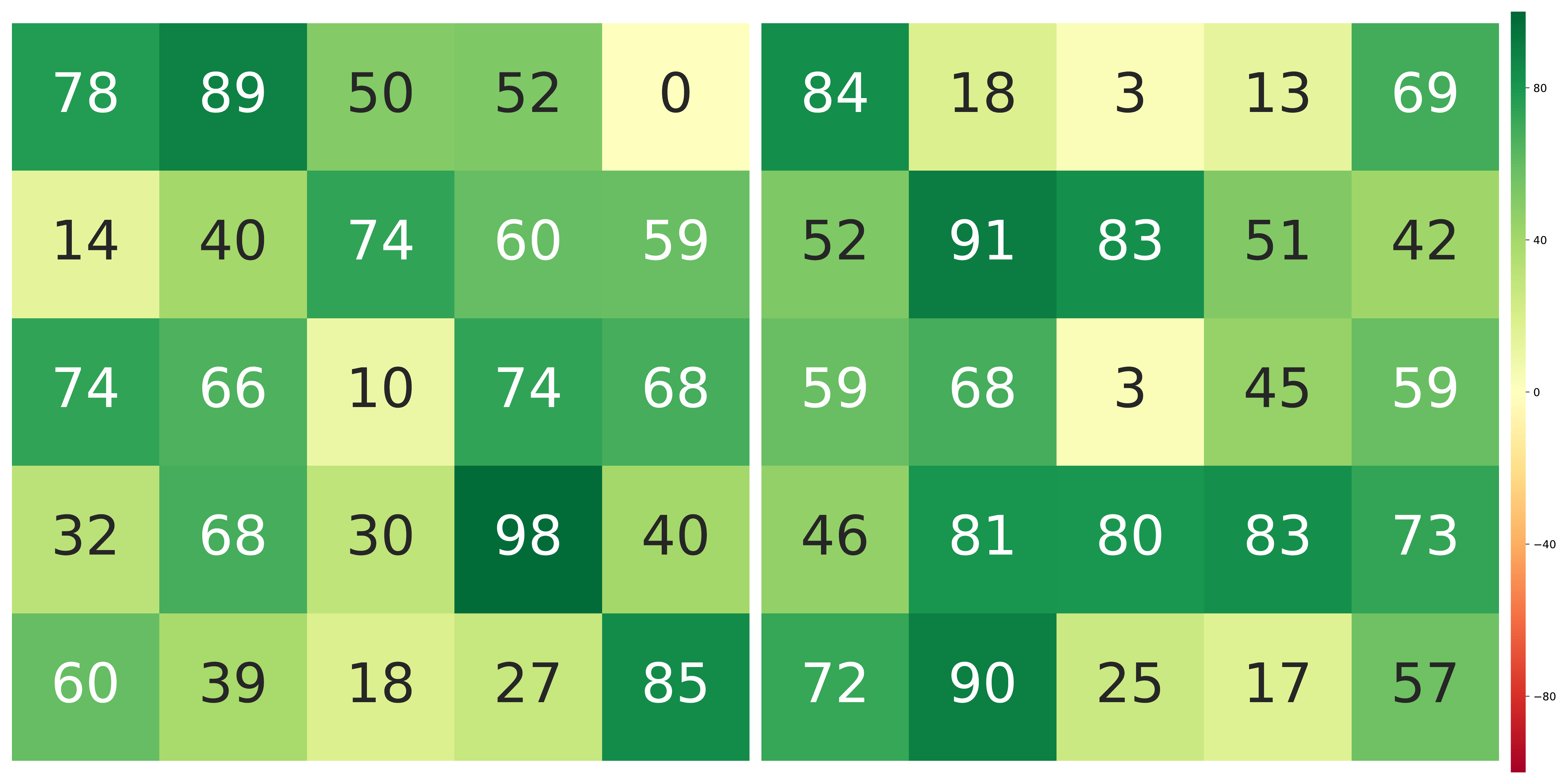}}
  \subfloat[Scenario after injecting 1 constraint in each utility function. Note that the colour of the constrained cells is not to scale to preserve differentiability of the other colours]{\includegraphics[width=0.5\columnwidth]{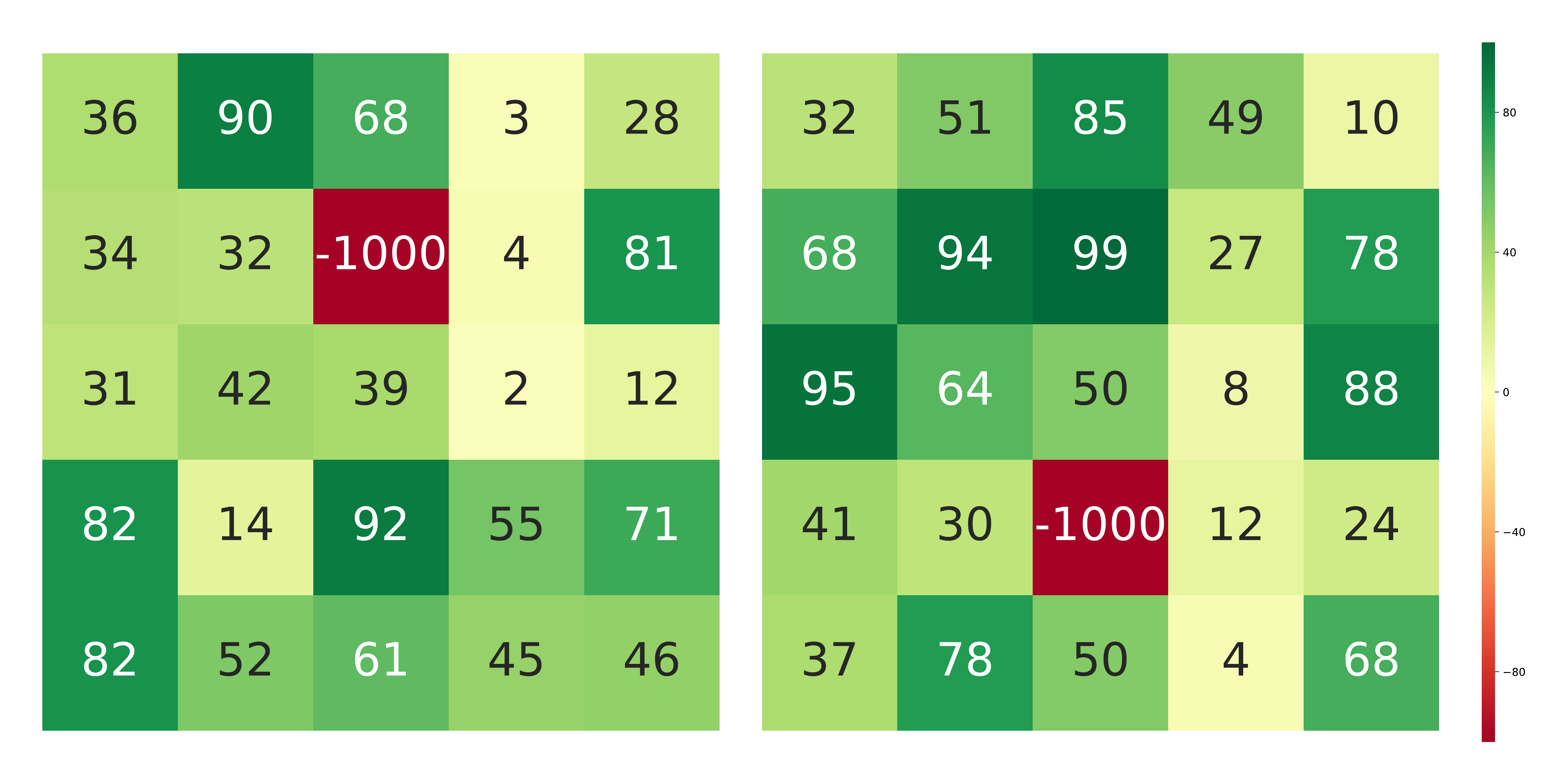}}
  \caption{Examples of generated utility matrices before and after injecting 1 constraint}
  \label{genExample}
\end{figure}

Initially 300 unique pairs of utility functions were generated by drawing from uniform distributions on either $\{0,1,\ldots,100\}$ or $\{0,1,\ldots,25\}$. The scenarios were drawn from two possible distributions to ensure that both sufficient impossible and possible configurations would be tested. Whether there are many, if any, mutually agreeable options in a configuration can be quite sensitive to randomness in the utility functions, and the reservation values the agents adopt, especially when the utility functions have a wide range.  For each of the 300 base scenarios, several variants were created by adding an equal number of constraints in both utility functions, up to a maximum of 12 per agent. Note that if we were to create a constraint in a value assignment where the opponent has very low utility, the constraint is unlikely to make a difference, since the opponent is not likely to make an offer that violates that constraint, meaning that the additional information doesn't get utilised. To avoid this problem we applied what we call \textit{constraint injection}. This means that if we want to introduce $n$ constraints in the utility function of agent $A$, we do this by determining the $n$ most favourable assignments for $B$ and overwrite the utilities for those assignments in $A$'s utility function with a value that is low enough to create a constraint. If $A$ has a maximum utility of $u_{max A}$ then a value lower than $ -u_{max A}$ is enough to ensure a constraint will be created. In this scenario, the theoretical best utility possible is 100. Therefore we used $-1000$ as our constraint value, to avoid potential boundary issues.  An example of a generated scenario sampled from $[0,100]$ before and after injecting 1 constraint in each utility function can be seen in Figure \ref{genExample}. 

We express reservation values as a percentage of the agent's maximum possible utility: an agent with a reservation value of $\frac 12$ will only accept offers that have at least half the utility of the best possible outcome. Firstly let $R_{lin} = \left\{\frac 12 + \frac{i}{20} | i \in \{0,1,\ldots,9\}\right\}$, i.e. 10 points spaced equally apart on $[\frac 12,1]$.  Furthermore let $R_{log} = \left\{10^{\log_{10}(\frac 12) - \frac{i\log_{10}(\frac 12)}{10}} | i \in \{0,1,\ldots,9\}\right\}$, i.e. 10 points in $[\frac 12,1]$ such that they are equally spaced in log-space. Pairs of reservation values were taken from either $(R_{lin})^2$ or $(R_{log})^2$.  Again, taking pairs from these two sets was to ensure that enough possible and impossible configurations would be explored.

\subsection{Running the simulations}
We introduced the two strategies used in this work---random sampling and concession---and how they work under AOP back in Section \ref{sec:aop}. We will now first explain how the agents adapt these strategies to function under ACOP.

The constraint-aware version of the random sampling agent will adjust the distribution it samples from, when a new constraint is introduced so that any assignment that has been ruled out is given probability 0. Since base random sampling agents construct offers by independently sampling from the possible values for each issue, an agents using ACOP can simply assign probability 0 to the values that were ruled out, and renormalise the distribution.  

The concession agent explores the negotiation space using breadth-first search with the utility function as a heuristic. When the constraint aware version of this agent receives a constraint, they adjust their utility function, but overwriting the utility of the value that is being ruled out by a value that is smaller than negative their best utility. This ensures that all offers not satisfying it will fall below the reservation value, ensuring that they will never generate an offer that violates a known constraint.

To summarise, for each of the configurations generated, as discussed in the last section, 4 simulations were run, corresponding to one of the following strategy and protocol pairs:
\begin{enumerate}
  \item Random sampling using AOP.
  \item Concession using AOP.
  \item Random sampling using ACOP.
  \item Concession using ACOP.
\end{enumerate}

To ensure that the negotiations would terminate, even if the configuration was impossible, a timeout of 400 rounds was introduced, meaning that each agent is allowed to make at most 200 offers. After this number of offers, agents would simply terminate the negotiation without reaching an agreement. In addition, the random sampling agent also terminates if it cannot discover an offer that is acceptable to themselves after 1000 samples, and the concession agent would terminate as soon as it cannot find new offers that have a utility above the reservation value. We chose these values as they were deemed to provide generous upper bounds for agents on the edge of a network. At the end of the negotiation three variables were collected:
\begin{enumerate}
  \item Whether the negotiation was successful.
  \item How many messages were exchanged during the entire negotiation.
  \item The utility achieved at the end of the negotiation by both agents.
\end{enumerate}
Here the utility achieved by each of the agents was equal to the utility of the offer that was accepted or $0$ if no agreement was reached.

\section{Results}
\label{sec:results}

\subsection{Impact of adopting ACOP on negotiation length}
\begin{figure}
  \centering
  \subfloat[Distribution of the length of the negotiations using concession]{\includegraphics[width=0.8\columnwidth]{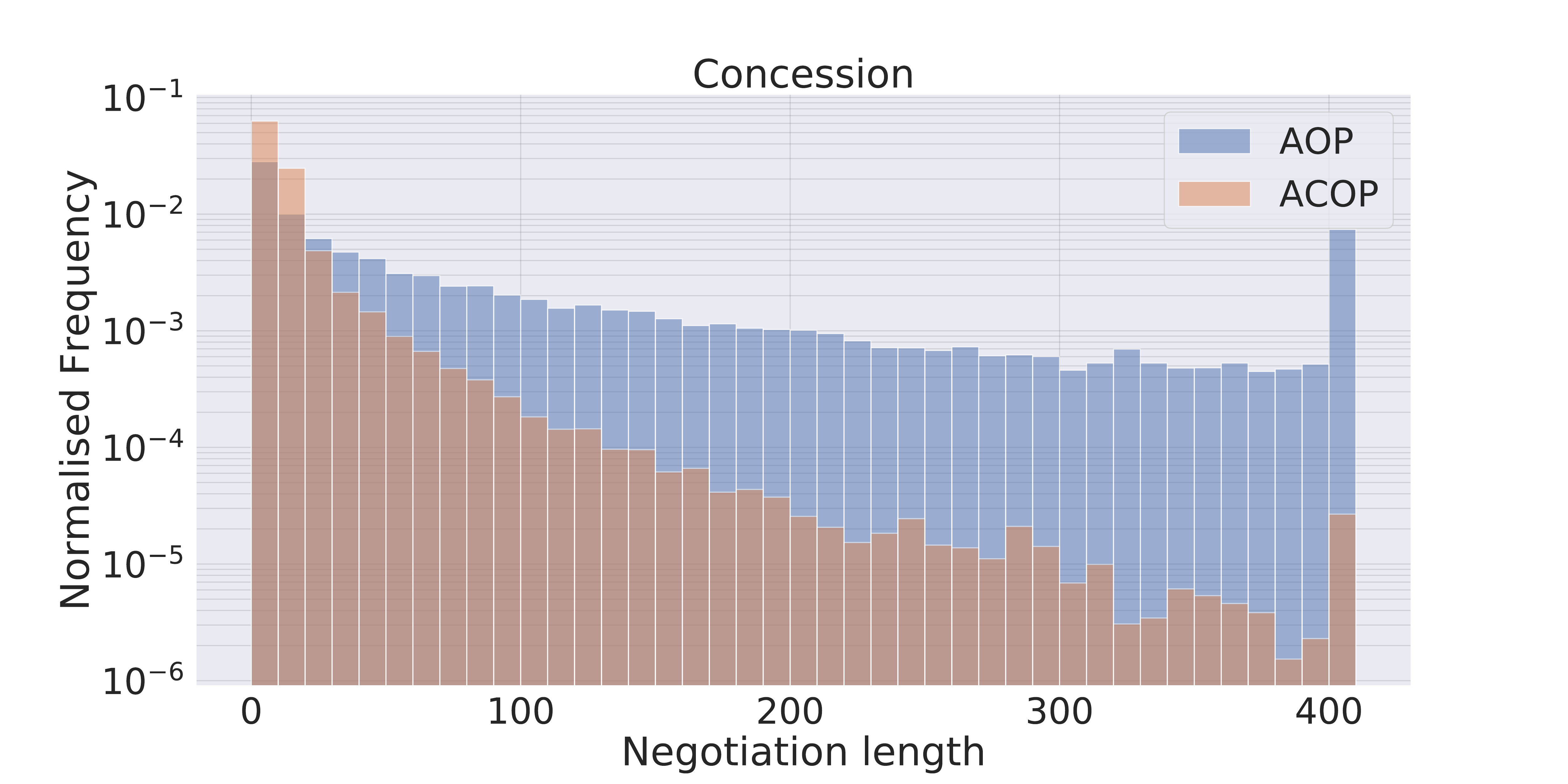}}
  
  \subfloat[Distribution of the length of the negotiations using random]{\includegraphics[width=0.8\columnwidth]{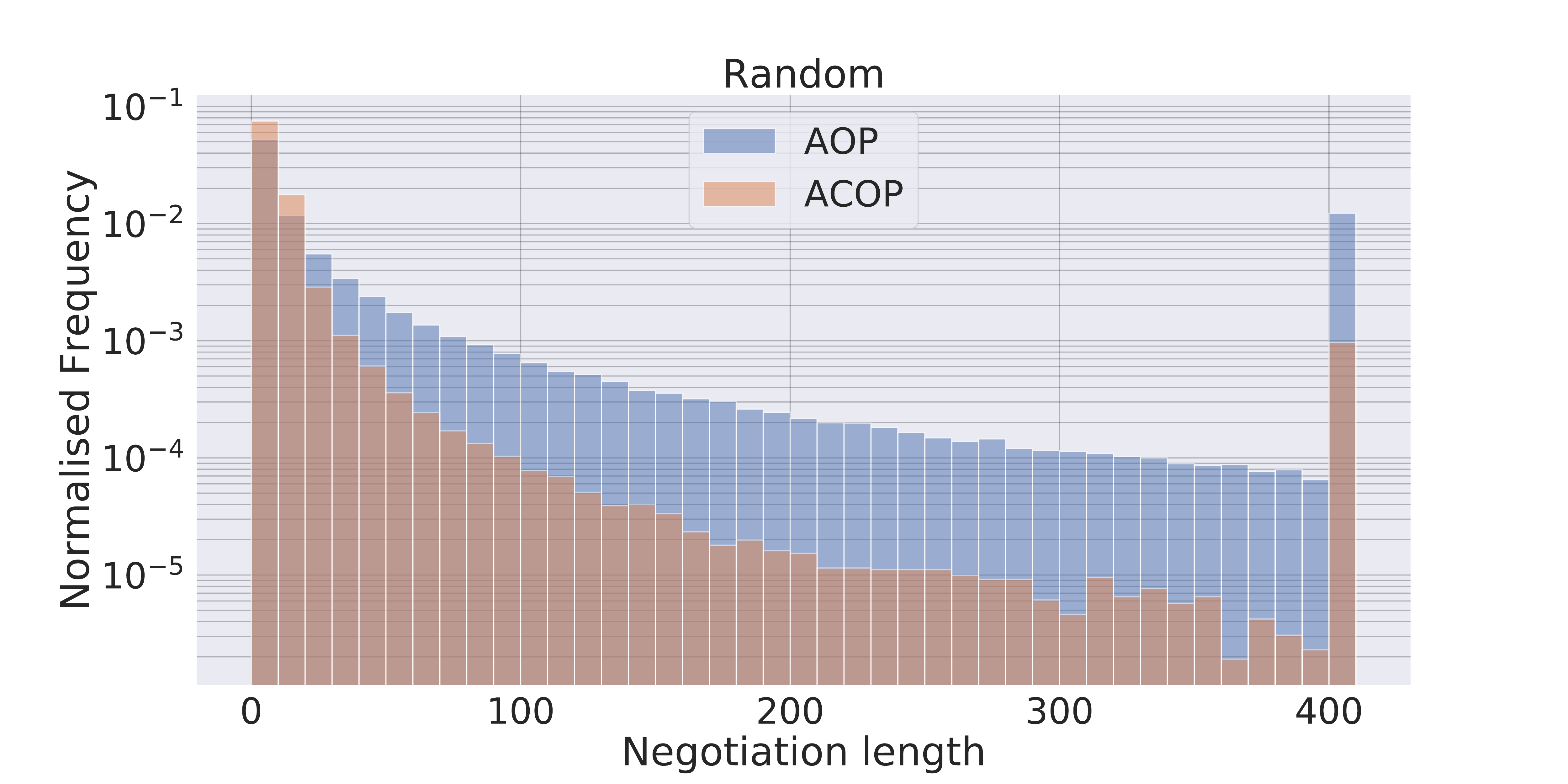}}
%   \subfloat[Negotiations using random, linear scale]{\includegraphics[width=\columnwidth]{len_freq_rand_lin.png}}
  \caption{A distribution plot of the length of all the negotiations simulated. Note the logarithmic scale. }
  \label{len_dists}
\end{figure}

 In this section, we study the impact that changing protocols, i.e., using constraints, has on negotiation length, keeping everything else fixed. Figure \ref{len_dists} plots for each strategy the frequency of different negotiation lengths, in a logarithmic scale. 

This shows that ACOP requires substantially fewer messages than AOP on average, evidenced by the fact that much more of the mass of the ACOP bars is concentrated near the left in both graphs. It is worth noting that the peak at the right of the graphs is mostly due to impossible negotiations. This solidifies the idea that no matter the `difficulty' of a negotiation, ACOP will on average terminate faster than AOP. We will investigate whether this means that ACOP achieves lower outcomes than AOP in the next section. 

\begin{figure}
    \centering
    \includegraphics[width=0.9\columnwidth]{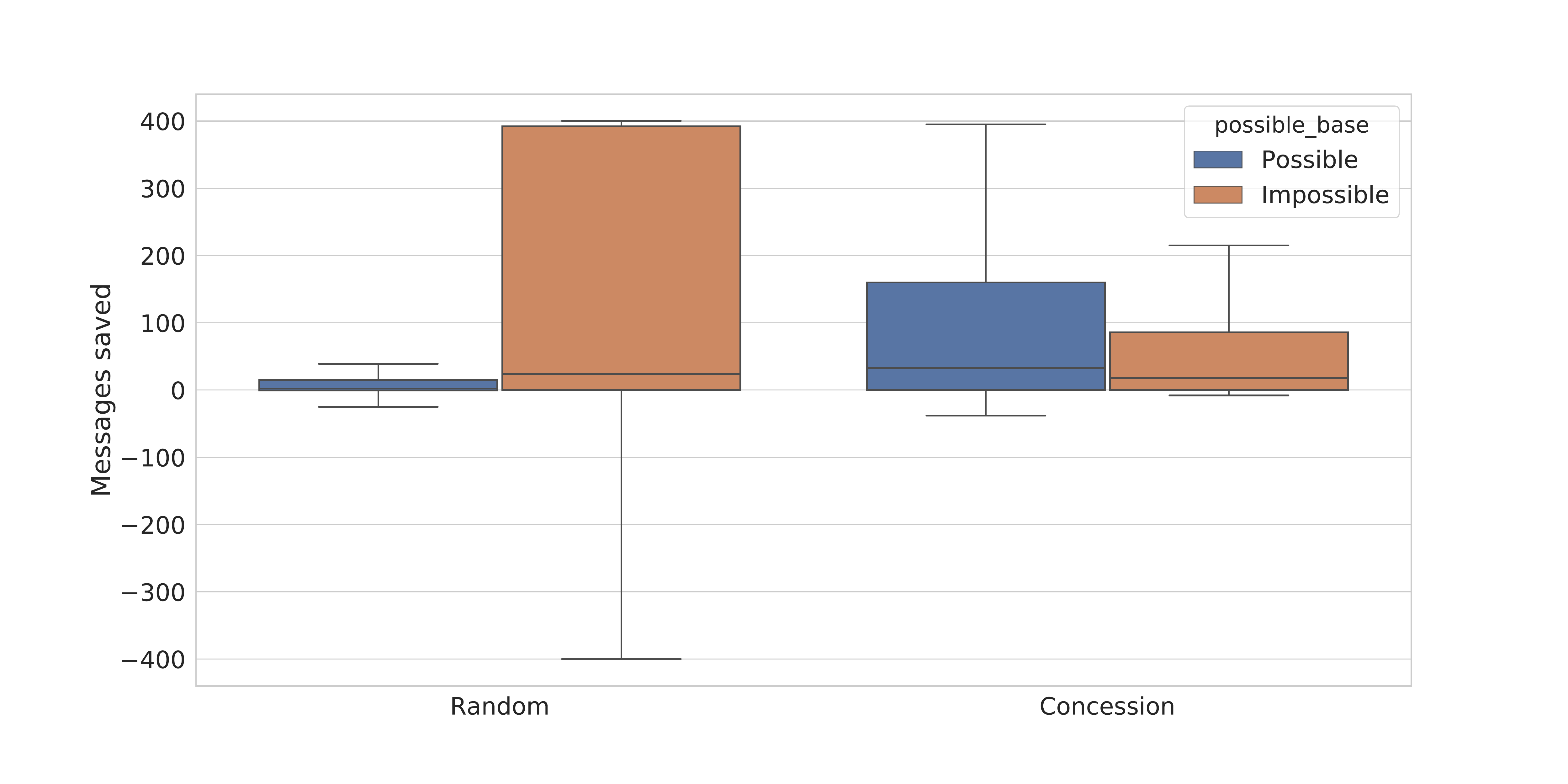}
    \caption{A box plot detailing the messages saved by using ACOP compared to AOP with identical configurations. }
    \label{len-improvement-box}
\end{figure}

We can get a more detailed understanding of the impact of using ACOP compared to AOP by looking at the box-plot in Figure \ref{len-improvement-box}. This figure depicts the number of messages saved by using ACOP instead of AOP in an identical configuration. Here we have broken down the data by two categories: The strategy used, and whether the configuration had a solution or not. 

For the agents using a random strategy, by far the most gains were made in the impossible configurations. Note that there are some configurations for which ACOP performed worse than AOP, as evidenced by the lower whisker. However, this is due to the randomness of the bidding. In these cases, the agents using AOP were simply unable to find an offer they found acceptable themselves, and thus terminated, while the constraints allowed the agents using ACOP to find proposals that were acceptable to themselves and thus kept negotiating. However we can deduce from the box plot that this is actually a relatively rare case. Even in cases where ACOP did not save a large number of messages, it almost never prolonged the negotiation by much if at all. 

For concession agents, ACOP saved more messages when the configurations did have a solution, meaning that ACOP allowed the concession agents to search the negotiation space much more effectively. In the case where the configurations were impossible, ACOP still decreased the number of messages used even if fewer messages were saved. This is due to the fact that a lot of the impossible negotiations still have large sets of offers that are acceptable to just one of the agents that have to be ruled out. When considering all simulations run, we see that ACOP saves an average of 75 messages and with a median of 8 messages saved. Considering that the distribution of negotiation lengths is heavily skewed towards the lower end, we consider this to be a very favourable result. With these observations we conclude that ACOP performs at least as well as AOP and improves upon AOP substantially in the majority of cases when considering the length of a negotiation.

\subsection{Impact of adopting ACOP on competitive advantage}
Before analysing the outcome of a negotiation in terms of utility two key observations need to be made. First of all, these results are highly dependent on the range of the utility functions. Secondly, the cost that agents incur by ending a negotiation without agreement can have a big impact on the results. The impact of having different non-agreement costs or very different utility functions is outside of the scope of this work. Therefore the agents in this work did not receive an additional penalty for failing to reach an agreement (i.e., a non-agreement was given utility 0 for both agents) and they were all given similar utility functions as discussed previously. 

Here we will investigate whether adopting ACOP negatively impacts the outcome of identical negotiations in which agents use AOP. To this end we compared the utility of the negotiations using ACOP to that of the negotiations of the same configuration but using AOP. In Table \ref{tab:util_ord} a per-strategy-breakdown can be seen of what percentage of the negotiations using ACOP had a much better, better, equal, worse or much worse outcome than negotiations of equal configurations using AOP. If ACOP has a higher utility, the configuration was classified as better. If ACOP had a utility of at least 10 higher (10\% of the theoretical maximum utility) it was classified as much better, with worse and much worse being defined similarly in the other direction. 

In this table, we can see that for the concession agent, the vast majority of negotiations using ACOP (81.68\%) had the exact same utility at the end as a negotiation of an identical configuration using AOP. While there were some cases in which ACOP performed slower, this happened in only roughly 3\% of all cases, and in only 0.55\% was the difference in utility bigger than 10. Conversely, in about 15\% of the cases ACOP achieved a higher utility at the end of a negotiation, and in roughly 9\% did it gain more than 10 utility above what AOP achieved. 

Looking at the percentages for the random agent, we see that while there are more negotiations where ACOP achieves a lower utility than AOP. This was to be expected, since agents will immediately accept any offer from the adversary they find acceptable. Furthermore, we can see that the frequencies are symmetrically distributed, meaning there are roughly equal numbers of configurations that achieved a higher utility using ACOP as there are configurations that achieved a lower utility using ACOP. This pattern can be easily explained by the randomness of the bidding of the agents.

\begin{table}[t]
    \centering
    \begin{tabular}{|l|l|r|}
\hline
       &            &  Percentage of total \\
Strategy & \makecell{ACOP compared to AOP} &                                     \\
\hline
Concession & Much better &                                9.06 \\
       & Better &                                5.88 \\
       & Equal &                               81.68 \\
       & Worse &                                2.82 \\
       & Much worse &                                0.55 \\\hline
Random & Much better &                                3.40 \\
       & Better &                               27.83 \\
       & Equal &                               38.44 \\
       & Worse &                               27.62 \\
       & Much worse &                                2.70 \\
\hline
\end{tabular}

    \caption{Table detailing the percentages of configurations broken down by strategy and how the utility of the outcome of ACOP compared to that of AOP. Configurations for which the difference in utility was greater than 10 would be classified as either much better (if ACOP did better), or much worse (if AOP did better)}
    \label{tab:util_ord}
\end{table}

With all of these observations, we conclude that using ACOP does not negatively affect the outcome of the negotiations in any systematic way. 

\section{Conclusion}
In this paper we proposed a novel extension to the Alternating Offers Protocol (AOP) called Alternating Constrained Offers Protocol (ACOP) which allows agents to express constraints to the adversary along with offering counter proposals. These constraints can be given to an agent apriori, or discovered using branch-and-bound algorithms. This protocol allows agents---especially agents deployed on low-power devices at the edge of a network---to terminate negotiations faster without consistently negatively impacting the utility of the outcome, allowing them to save bandwidth without the need to equip them with sophisticated reasoning capabilities. We explored the impact that this extension has on the length of the negotiations as well as on the utility achieved at the end of the negotiation. We empirically showed that this extension substantially reduces the number of messages agents have to exchange during a negotiation. When agreement is possible, using ACOP helps agents to come to an agreement faster, and when agreement is impossible, agents using ACOP terminate much faster than agents using AOP both when agents adopt a probabilistic or a deterministic search method. In addition, we showed that using ACOP has no systematic negative impact on the quality of the outcome in terms of utility when compared to the same strategies using AOP. 

While the results of this work were promising, the scenarios and strategies used to produce them were not very complex. Future work will include investigating the performance of ACOP under non-linear utility functions, and with more sophisticated strategies and opponent models, comparing also with other approaches for dealing for instance with fuzzy constraints \cite{luoFuzzyConstraintBased2003}, and with also much larger large, non-linear agreement spaces \cite{Jonge2015}. Another avenue will be to understand the impact of using soft constraints rather than hard ones.

\bibliographystyle{splncs04}
\bibliography{sams_library,references}  % put name of your .bib file here

\begin{thebibliography}{10}
\providecommand{\url}[1]{\texttt{#1}}
\providecommand{\urlprefix}{URL }
\providecommand{\doi}[1]{https://doi.org/#1}

\bibitem{Aydogan2017}
Aydo{\u g}an, R., Festen, D., Hindriks, K.V., Jonker, C.M.: Alternating
  {{Offers Protocols}} for {{Multilateral Negotiation}}. In: Modern
  {{Approaches}} to {{Agent}}-Based {{Complex Automated Negotiation}}, pp.
  153--167. {Springer International Publishing}, {Cham} (2017).
  \doi{10.1007/978-3-319-51563-2\_10}

\bibitem{azarNonCooperativeFrameworkCoordinating2019}
Azar, A.G., Nazaripouya, H., Khaki, B., Chu, C., Gadh, R., Jacobsen, R.H.: A
  {{Non}}-{{Cooperative Framework}} for {{Coordinating}} a {{Neighborhood}} of
  {{Distributed Prosumers}}. IEEE Transactions on Industrial Informatics
  \textbf{15}(5),  2523--2534 (May 2019). \doi{10.1109/TII.2018.2867748}

\bibitem{baarslagLearningOpponentAutomated2016}
Baarslag, T., Hendrikx, M.J.C., Hindriks, K.V., Jonker, C.M.: Learning about
  the opponent in automated bilateral negotiation: A comprehensive survey of
  opponent modeling techniques. Autonomous Agents and Multi-Agent Systems
  \textbf{30}(5),  849--898 (Sep 2016). \doi{10.1007/s10458-015-9309-1}

\bibitem{baarslagAcceptingOptimallyAutomated2013a}
Baarslag, T., Hindriks, K.V.: Accepting {{Optimally}} in {{Automated
  Negotiation}} with {{Incomplete Information}}. In: Proceedings of the 2013
  {{International Conference}} on {{Autonomous Agents}} and {{Multi}}-Agent
  {{Systems}}. pp. 715--722. {{AAMAS}} '13, {International Foundation for
  Autonomous Agents and Multiagent Systems}, {Richland, SC} (2013)

\bibitem{fatima2014principles}
Fatima, S., Kraus, S., Wooldridge, M.: Principles of automated negotiation.
  Cambridge University Press (2014)

\bibitem{gaciarzConstraintBasedNegotiationModel2015}
Gaciarz, M., Aknine, S., Bhouri, N.: Constraint-{{Based Negotiation Model}} for
  {{Traffic Regulation}}. In: 2015 {{IEEE}}/{{WIC}}/{{ACM International
  Conference}} on {{Web Intelligence}} and {{Intelligent Agent Technology}}
  ({{WI}}-{{IAT}}). vol.~2, pp. 320--327. {IEEE}, {Singapore} (Dec 2015).
  \doi{10.1109/WI-IAT.2015.72}

\bibitem{godeAllocativeEfficiencyMarkets1993}
Gode, D.K., Sunder, S.: Allocative {{Efficiency}} of {{Markets}} with
  {{Zero}}-{{Intelligence Traders}}: {{Market}} as a {{Partial Substitute}} for
  {{Individual Rationality}}. Journal of Political Economy  \textbf{101}(1),
  119--137 (Feb 1993). \doi{10.1086/261868}

\bibitem{hindriksUsingOpponentModels2009}
Hindriks, K., Jonker, C., Tykhonov, D.: Using {{Opponent Models}} for
  {{Efficient Negotiation}}. In: Proceedings of {{The}} 8th {{International
  Conference}} on {{Autonomous Agents}} and {{Multiagent Systems}} - {{Volume}}
  2. pp. 1243--1244. {{AAMAS}} '09, {International Foundation for Autonomous
  Agents and Multiagent Systems}, {Richland, SC} (2009)

\bibitem{delaHoz2015}
de~la Hoz, E., Gimenez-Guzman, J.M., Marsa-Maestre, I., Orden, D.: Automated
  negotiation for resource assignment in wireless surveillance sensor networks.
  Sensors (Basel, Switzerland)  \textbf{15}(11),  29547--29568 (Nov 2015).
  \doi{10.3390/s151129547}, \url{https://www.ncbi.nlm.nih.gov/pubmed/26610512},
  26610512[pmid]

\bibitem{Jonge2015}
Jonge, D., Sierra, C.: $\hbox{NB}^{3}$: A multilateral negotiation algorithm
  for large, non-linear agreement spaces with limited time. Autonomous Agents
  and Multi-Agent Systems  \textbf{29}(5),  896–942 (Sep 2015).
  \doi{10.1007/s10458-014-9271-3},
  \url{https://doi.org/10.1007/s10458-014-9271-3}

\bibitem{klein2003negotiating}
Klein, M., Faratin, P., Sayama, H., Bar-Yam, Y.: Negotiating complex contracts.
  Group Decision and Negotiation  \textbf{12}(2),  111--125 (2003)

\bibitem{45648}
Kone\v{c}n\'y, J., McMahan, H.B., Yu, F.X., Richtarik, P., Suresh, A.T., Bacon,
  D.: Federated learning: Strategies for improving communication efficiency.
  In: NIPS Workshop on Private Multi-Party Machine Learning (2016),
  \url{https://arxiv.org/abs/1610.05492}

\bibitem{lamparterAgentbasedMarketPlatform2010}
Lamparter, S., Becher, S., Fischer, J.G.: An {{Agent}}-based {{Market
  Platform}} for {{Smart Grids}}. In: Proceedings of the 9th {{International
  Conference}} on {{Autonomous Agents}} and {{Multiagent Systems}}: {{Industry
  Track}}. pp. 1689--1696. {{AAMAS}} '10, {International Foundation for
  Autonomous Agents and Multiagent Systems}, {Richland, SC} (2010)

\bibitem{lopesConcessionStrategiesNegotiating2012a}
Lopes, F., Coelho, H.: Concession {{Strategies}} for {{Negotiating Bilateral
  Contracts}} in {{Multi}}-agent {{Electricity Markets}}. In: 2012 23rd
  {{International Workshop}} on {{Database}} and {{Expert Systems
  Applications}}. pp. 321--325 (Sep 2012). \doi{10.1109/DEXA.2012.24}

\bibitem{luoFuzzyConstraintBased2003}
Luo, X., Jennings, N.R., Shadbolt, N., Leung, H.f., Lee, J.H.m.: A fuzzy
  constraint based model for bilateral, multi-issue negotiations in
  semi-competitive environments. Artificial Intelligence  \textbf{148}(1),
  53--102 (Aug 2003). \doi{10.1016/S0004-3702(03)00041-9}

\bibitem{wangIntelligentNegotiationAgent2012}
Wang, Z., Wang, L.: Intelligent negotiation agent with learning capability for
  energy trading between building and utility grid. In: {{IEEE PES Innovative
  Smart Grid Technologies}}. pp.~1--6 (May 2012).
  \doi{10.1109/ISGT-Asia.2012.6303167}

\bibitem{Xu:2019:EFL:3321408.3323080}
Xu, Z., Li, L., Zou, W.: Exploring federated learning on battery-powered
  devices. In: Proceedings of the ACM Turing Celebration Conference - China.
  pp. 6:1--6:6. ACM TURC '19, ACM, New York, NY, USA (2019).
  \doi{10.1145/3321408.3323080},
  \url{http://doi.acm.org/10.1145/3321408.3323080}

\bibitem{8462804}
{Yang}, S., {Peng}, D., {Meng}, T., {Wu}, F., {Chen}, G., {Tang}, S., {Li}, Z.,
  {Luo}, T.: On designing distributed auction mechanisms for wireless spectrum
  allocation. IEEE Transactions on Mobile Computing  \textbf{18}(9),
  2129--2146 (Sep 2019). \doi{10.1109/TMC.2018.2869863}

\bibitem{yangMultidemandNegotiationModel2019}
Yang, Y., Luo, X.: A {{Multi}}-demand {{Negotiation Model}} with {{Fuzzy
  Concession Strategies}}. In: Rutkowski, L., Scherer, R., Korytkowski, M.,
  Pedrycz, W., Tadeusiewicz, R., Zurada, J.M. (eds.) Artificial
  {{Intelligence}} and {{Soft Computing}}. pp. 689--707. Lecture {{Notes}} in
  {{Computer Science}}, {Springer International Publishing} (2019)

\bibitem{yassineDoubleAuctionMechanisms2019}
Yassine, A., Hossain, M.S., Muhammad, G., Guizani, M.: Double auction
  mechanisms for dynamic autonomous electric vehicles energy trading. IEEE
  Transactions on Vehicular Technology  \textbf{68}(8),  7466--7476 (Aug 2019).
  \doi{10.1109/TVT.2019.2920531}

\bibitem{Zhang2010}
Zhang, G., Jiang, G.R., Huang, T.Y.: Design of argumentation-based multi-agent
  negotiation system oriented to {{E}}-commerce. International Conference on
  Internet Technology and Applications, ITAP 2010 - Proceedings pp.~1--6
  (2010). \doi{10.1109/ITAPP.2010.5566198}

\end{thebibliography}
\end{document}